\documentclass[letterpaper]{article} 
\usepackage{aaai25}  
\usepackage{times}  
\usepackage{helvet}  
\usepackage{courier}  
\usepackage[hyphens]{url}  
\usepackage{graphicx} 
\usepackage{natbib}  
\usepackage{caption} 
\usepackage{algorithm}
\usepackage{algorithmic}
\usepackage{multirow}
\usepackage{makecell}
\usepackage{colortbl}

\usepackage{newfloat}
\usepackage{listings}
\usepackage{amsmath,amssymb,amsfonts}
\DeclareCaptionStyle{ruled}{labelfont=normalfont,labelsep=colon,strut=off} 
\floatstyle{ruled}
\newfloat{listing}{tb}{lst}{}
\floatname{listing}{Listing}
\title{Universal Medical Image Representation Learning with Compositional Decoders}
\author{
    Kaini Wang\textsuperscript{\rm 1}\equalcontrib ,
    Ling Yang\textsuperscript{\rm 2}\equalcontrib,
    Siping Zhou\textsuperscript{\rm 1} ,
    Guangquan Zhou \textsuperscript{\rm 1}\thanks{Corresponding author.},
    Wentao Zhang \textsuperscript{\rm 2},
    Bin CUI \textsuperscript{\rm 2},
    Shuo Li \textsuperscript{\rm 3}
}
\affiliations{
    \textsuperscript{\rm 1} Southeast University, China\\

   \textsuperscript{\rm 2} Peking University, China\\
   \textsuperscript{\rm 3} Case Western Reserve University, USA\\
    
}

\usepackage{bibentry}

\begin{document}
	
	\maketitle
	
	\begin{abstract}
		Visual-language models have advanced the development of universal models, yet their application in medical imaging remains constrained by specific functional requirements and the limited data. Current general-purpose models are typically designed with task-specific branches and heads, which restricts the shared feature space and the flexibility of model. To address these challenges, we have developed a decomposed-composed universal medical imaging paradigm (UniMed) that supports tasks at all levels. To this end, we first propose a decomposed decoder that can predict two types of outputs- pixel and semantic, based on a defined input queue. Additionally, we introduce a composed decoder that unifies the input and output spaces and standardizes task annotations across different levels into a discrete token format. The coupled design of these two components enables the model to flexibly combine tasks and mutual benefits. Moreover, our joint representation learning strategy skilfully leverages large amounts of unlabeled data and unsupervised loss, achieving efficient one-stage pretraining for more robust performance. Experimental results show that UniMed achieves state-of-the-art performance on eight datasets across all three tasks and exhibits strong zero-shot and 100-shot transferability. We will release the code and trained models upon the paper's acceptance.
	\end{abstract}
	
	%
	
	\section{Introduction}
	\label{sec:intro}
	
	\begin{figure}[tb]
		\centering
		\includegraphics[width=0.9\columnwidth]{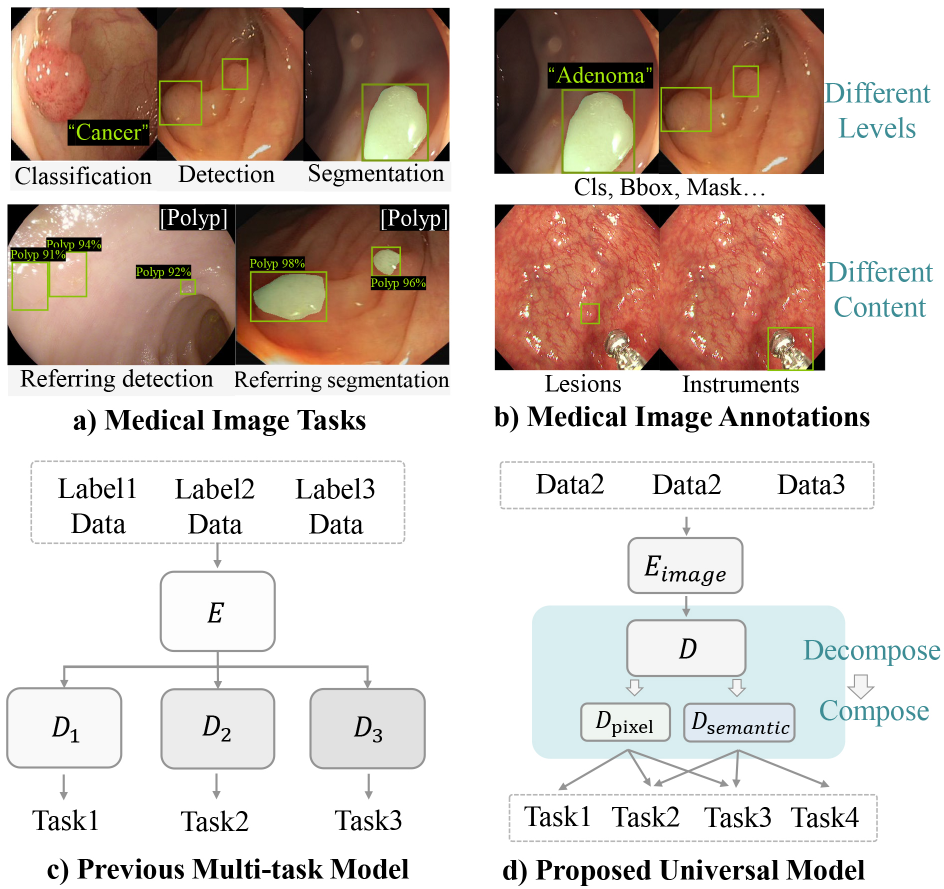}
		\caption{ a) The broad range of tasks in medical image analysis. b) The diversity of annotations both across tasks and between different datasets. c) Existing models require task-specific branches or heads. d) The proposed universal model seamlessly supports all levels of tasks by matching the decompose output decoder with the compose label decoder.
		}
		\label{fig1}
	\end{figure}
	
	Vision-language models have significant success in establishing a universal framework that not only reduces the cost of processing different tasks but also supports collaboration among them \cite{radford2021learning, alayrac2022flamingo}. However, for a universal model in medical image analysis to be viable in real clinical settings \cite{moor2023foundation}, it requires 1) the versatility to simultaneously handle semantic understanding and visual tasks (e.g., not only locating lesions but also identifying their types). 2) the seamless transition between different tasks, allowing users to tailor functionalities based on the specific scenario (e.g., toggle detection and segmentation tasks according to whether they involve lesion screening or resection procedures). 3) the robust transferability, ensuring the model's adeptness in delivering high-quality predictions even when confronted with new data.
	
	The performance of universal models is largely driven by increasingly complete data, which imposes great demands on data construction and maintenance \cite{liu2023clip}. However, the scale of data available in medical imaging is relatively limited compared to natural images due to the high cost and expertise required for data collection and annotation. Consequently, current research focuses on optimizing models on fixed data \cite{wang2022awsnet, zhou2023dsanet}. Such specialized frameworks often encounter sudden performance drops when applied to other datasets or tasks. In reality, we observe that for any modality of medical imaging (such as endoscopic imaging), a large amount of data can be aggregated from existing public datasets. However, the diversity among these datasets hinders their direct integration and use. Therefore, this work emphasizes that the key is to maximize the effectiveness of existing medical image data in developing universal models.
	
	One challenge in designing such a model is the different levels of annotations across tasks. For example (Fig. \ref{fig1} a b), the classification task involves semantic annotation at the image level, the segmentation task requires pixel-level annotation, and the referring segmentation task combines text and pixel-level annotations. Until recently, attempts have been made to develop multi-task learning models \cite{qin2022medical, wan2024med}, which have demonstrated encouraging cross-task generalization capabilities. However, most of these studies involve additional branches or heads, leading to increased model complexity and difficulties in balancing tasks. Furthermore, the unification of all levels of tasks in medical images—whether at the image, region, or pixel level—has yet to be fully achieved. 
	
	Another challenge is the different content of annotations among datasets (Fig. \ref{fig1} b)). Unlike fully labeled benchmarks in natural image analysis, some medical imaging datasets contain annotations solely for lesions, while others provide annotations for instruments. Mainstream methods \cite{yu2019crossbar, isensee2021nnu} address this issue by splitting differently labeled datasets into several subsets and training the network on each subset to complete specific tasks. While this strategy is intuitive, it significantly increases computational complexity. Additionally, this design limits the sharing of knowledge across different annotations, leaving the common semantic space for task understanding largely unexplored.
	
	Based on the above observations, we propose a universal medical image analysis model (UniMed) capable of performing various medical imaging tasks at all levels, including image, region, an pixel levels (Fig. \ref{fig1} d)). Specifically, to reconcile the diversity in annotations across different tasks, we introduce a composed decoder that standardizes annotations into a discrete label format. We design an decomposed decoder that, instead of dividing the output by task, decomposes all tasks into pixel-level and semantic-level components. The innovative coupling design of the separable decoder and label converter unifies the input and output space, enabling flexible model combinations to support various task interactions seamlessly. Additionally, UniMed is equipped with an annotation understanding branch at the input stage to encode nouns and texts in the task, promoting the learning of a shared visual semantic space to accommodate the inherent diversity of tasks. Furthermore, to leverage large amounts of unlabeled data, we propose a joint representation learning strategy that enables unlabeled data to guide the encoder in extracting effective representations through contrastive learning. The key differentiator of our approach is its end-to-end framework without supplementary branches and modules. 
	
	Our main contributions are summarized as the following:
	\begin{itemize}
		\item \textbf{A new universal medical image paradigm.} Different from the traditional vision-language approach, we are the first to start from the limitations of data and propose a universal paradigm adaptable to diverse tasks and datasets for medical image learning. 
		
		\item \textbf{A decomposed-composed way for multi-tasking.} We propose a unified separable decoder that formulates all tasks into unified processing at the pixel level and semantic level. Combination with text sequences on the input side can support all tasks and promote cross-task collaborative development.
		
		\item \textbf{An efficient algorithm for representation learning.} We propose an effective learning strategy that enables full utilize unlabeled data through comparative learning, and jointly improve the transferability of the model with labeled data.
		
		\item \textbf{Significant experimental improvements.} The proposed model demonstrates strong zero-shot and 100-shot capabilities on eight datasets for three medical tasks and exceeds the current state-of-the-art specialized and generalist methods after fine-tuning.
		
	\end{itemize}
	
	\section{Related Work}
	
	\subsection{Medical Image Analysis Tasks}
	In the field of medical image analysis, three critical tasks are predominantly recognized: lesion detection, classification, and segmentation \cite{chen2022recent}.  Detection entails identifying the location of a lesion within an image based on a textual query \cite{tiu2022expert}, which is crucial for clinical finding and localization of abnormalities. Lesion classification involves assigning a class label to an image \cite{wang2022ffcnet} or a specified target region \cite{murtaza2020deep}. Segmentation requires generating pixel-level labels for an entire image \cite{ronneberger2015u}, aiding in the clinical demarcation of lesion boundaries. Multi-task models for the aforementioned tasks typically use a shared visual backbone to produce visual embeddings \cite{liu2023clip}, followed by individual branches tailored for each specific task. While these task-specific learning frameworks are effective for particular datasets, they lack generality and necessitate designing from scratch for new tasks.

	\subsection{Medical Universal Models}
	The emergence of large-scale models has significantly revolutionized the field of medical image analysis \cite{rajpurkar2022ai}. Recent studies have been increasingly focused on developing general-purpose medical artificial intelligence (AI) models \cite{moor2023foundation}. A notable trend is the incorporation of the Segment Anything Model (SAM) \cite{wu2023medical,ma2023segment,cheng2023sam,zhang2023customized}, which amalgamates medical domain knowledge for medical image segmentation and their application across various segmentation tasks. Concurrently, the emergence of image-text models has garnered considerable attention \cite{liu2023clip,ye2023uniseg,zhao2023one,wang2022medclip}. These models interpret image features through task-specific prompts, encompassing diverse modalities and domains, and represent a stride toward prompt-driven universal models. Most current methodologies predominantly concentrate on medical image segmentation \cite{butoi2023universeg} without adequately acknowledging the interconnectedness and uniformity across various medical tasks. Our research aims to bridge this gap by proposing a unified approach capable of handling all three tasks simultaneously, enabling training with diverse annotations and tasks, and building the foundation for more versatile and universally applicable medical image analysis.
	
	\begin{figure*}[tb]
		\centering
		\includegraphics[width=0.85\textwidth]{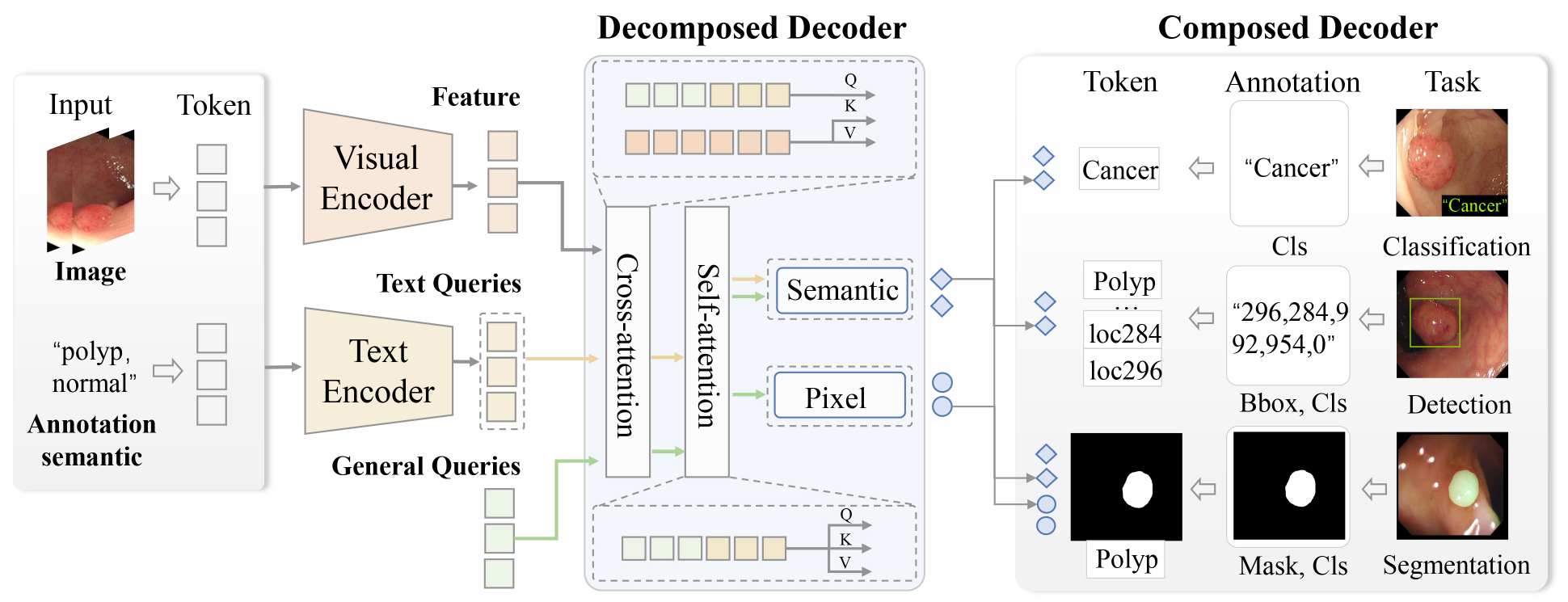}
		\caption{Overview of UniMed, consisting of four core components: a visual encoder, a text encoder, and a decomposed decoder and composed decoders. The decomposed decoder serves to amalgamate the output space of tasks into discrete tokens, encapsulating both semantic and pixel outputs. Similarly, composed decoders are harmonized into the same formats via a label converter to support cross-task learning. 
		}
		\label{fig2}
	\end{figure*}
	

	\section{Method}
	This paper establish a universal model capable of simultaneously handling various medical tasks, enabling concurrent learning from diverse labeled and unlabeled data sources without the need for task-specific parameters. The UniMed (Fig. \ref{fig2}) contains several core components: a universal vision-language architecture, a decomposed decoder for task output, a composed decoder with unified annotations, and a data-efficient joint training strategy. Such architecture enables the utilization of all annotated medical image types, promoting knowledge sharing across tasks, facilitating representation on labeled and unlabeled datasets, and benefiting many different downstream applications. 
	
	\subsection{Universal Medical Vision-Language Architecture}
	UniMed comprises a visual-language encoder and a dual-output decoder architecture (Fig. \ref{fig2}). Features are learned through visual and textual encoders upon receiving an input image. Subsequently, guided by the task labels, a decoder is employed to autoregressively predict the sequence.
	
	\textbf{Visual Encoder.} To standardize the input space into discrete tokens, the encoder must be transformer-based. Therefore, the visual encoder  $\texttt{Enc}_v$ utilizes the Swin-Transformer as its backbone, given its widely proven effectiveness. Given an input image $I$, this component extracts its layer features $V_l$ to derive the final multi-scale visual feature $V$ representation:
	\begin{equation}\label{eq1}
		V = \texttt{Enc}_v(I) = [V_1,V_2,...,V_L]
	\end{equation} 
	where L is the number of layers.
	
	\textbf{Text Encoder.} The text encoder is designed to capture annotation semantics and learn a broad spectrum of corpus knowledge. Since the specific annotations vary from dataset to dataset, the focus may be on lesions, instruments, or multiple annotation types, etc. For a piece of text $T$ generated from annotations, SentencePiece \cite{kudo2018sentencepiece} is first employed to divide the words and convert them into discrete token sequences. The text encoder, $\texttt{Enc}_t$, consists of multiple layers of Transformers that process the input text sequence. This process forms a text input queue $Q_t$, as follows:
	\begin{equation}\label{eq2}
		Q_t = \texttt{Enc}_t(T) = [q_{t}^{1},q_{t}^{2},...,q_{t}^{n}]
	\end{equation} 
	where $n$ is the length of the query. 
	
	\textbf{Decomposed Decoder.} The input to the model includes visual features $V$, a text queue $Q_t$, and a general queue $Q_g = [q_{g}^{1},q_{g}^{2},...,q_{g}^{m}]$, while the output consists of a pixel  $O_p$ and a semantic $O_s$. The flexible combination of these inputs and outputs is capable of supporting both general and referring tasks. The decomposed decoder is composed of stacked Transformers. It initially captures the global features of the image by computing the masked cross-attention $\texttt{Att}_{cross}$ among the three inputs \cite{cheng2022masked} and a self-attention  $\texttt{Att}_{self}$ mechanism to generate the queue for the subsequent layer:
	\begin{equation}\label{eq3}
		[\hat{Q}_t^{l}, \hat{Q}_g^{l}] = \texttt{Att}_{cross}([Q_t^{l-1}, Q_g^{l-1}], V)
	\end{equation} 
	\begin{equation}\label{eq4}
		[Q_t^{l}, Q_g^{l}] = \texttt{Att}_{self}([\hat{Q}_t^{l-1}, \hat{Q}_g^{l-1}])
	\end{equation}
	where $l$ represents the $l \texttt{-th}$ layer. For general tasks, the last general query  $Q_g$ is utilized as the global image representation. For referring tasks, the text query $Q_t$  serves as the referring feature, while the general query $Q_g$ is combined with it to produce the final representation. 
	
	For pixel-level output, the decoder utilizes global image features from the general queue to produce output $O_p = [ O_{p}^{1}, O_{p}^{2},..., O_{p}^{m}] $, facilitating a nuanced understanding of the image at a fine-grained level. Moreover, for semantic-level output, the decoder relies on both the general and text queues $O_s = [O_{s}^{1}, O_{s}^{2},..., O_{s}^{m+n}] $ to facilitate higher-level semantic understanding and generation.
	
	\textbf{Overall operation.} UniMeds's encoder encompasses the visual $\texttt{Enc}_v$ and text $\texttt{Enc}_t$ feature extraction branch, whereas the decomposed decoder $\texttt{Dec}$ utilizes visual features $V$, textual queues $Q_t$, and general $Q_g$ queues to forecast both pixel-level $O_p$ and semantic-level $O_s$ outputs \cite{cheng2022masked}. The overall operation can be expressed as:
	\begin{equation}\label{eq5}
		[O_p, O_s] = \texttt{Dec} (V, (Q_t, Q_g)) 
	\end{equation} 
	
	\subsection{Composed Decoders} \label{sec3.2}
	Since the decomposed decoder represents the output in terms of semantic and pixel outputs, the composed decoder unifies the labels of different tasks into a format that can be expressed through these two outputs.
	
	\textbf{Unified annotations.} For classification, the model outputs the image category, with the corresponding annotation being the category text. We match the text with the semantic output path by tokenizing it using SentencePiece. For detection,  the model identifies the location of the target area, with the annotation being the diagonal coordinates of the bounding box. To encode this sparse structure, we encode the sparse structure by expanding the vocabulary with 1000 special tokens \cite{chen2021pix2seq}. The bounding box is then represented by four tokens, two indicating the upper-left corner and the other two representing the lower-right corner, together with the category, serve as semantic outputs. For segmentation,  the model generates a result for each pixel, with the label being the pixel-level mask. The category annotation is processed using the method described earlier. The label image is encoded into discrete tokens, enabling the simultaneous generation of both pixel-level and semantic-level predictions.
	
\begin{figure}[tb]
		\centering
		\includegraphics[width=0.9\columnwidth]{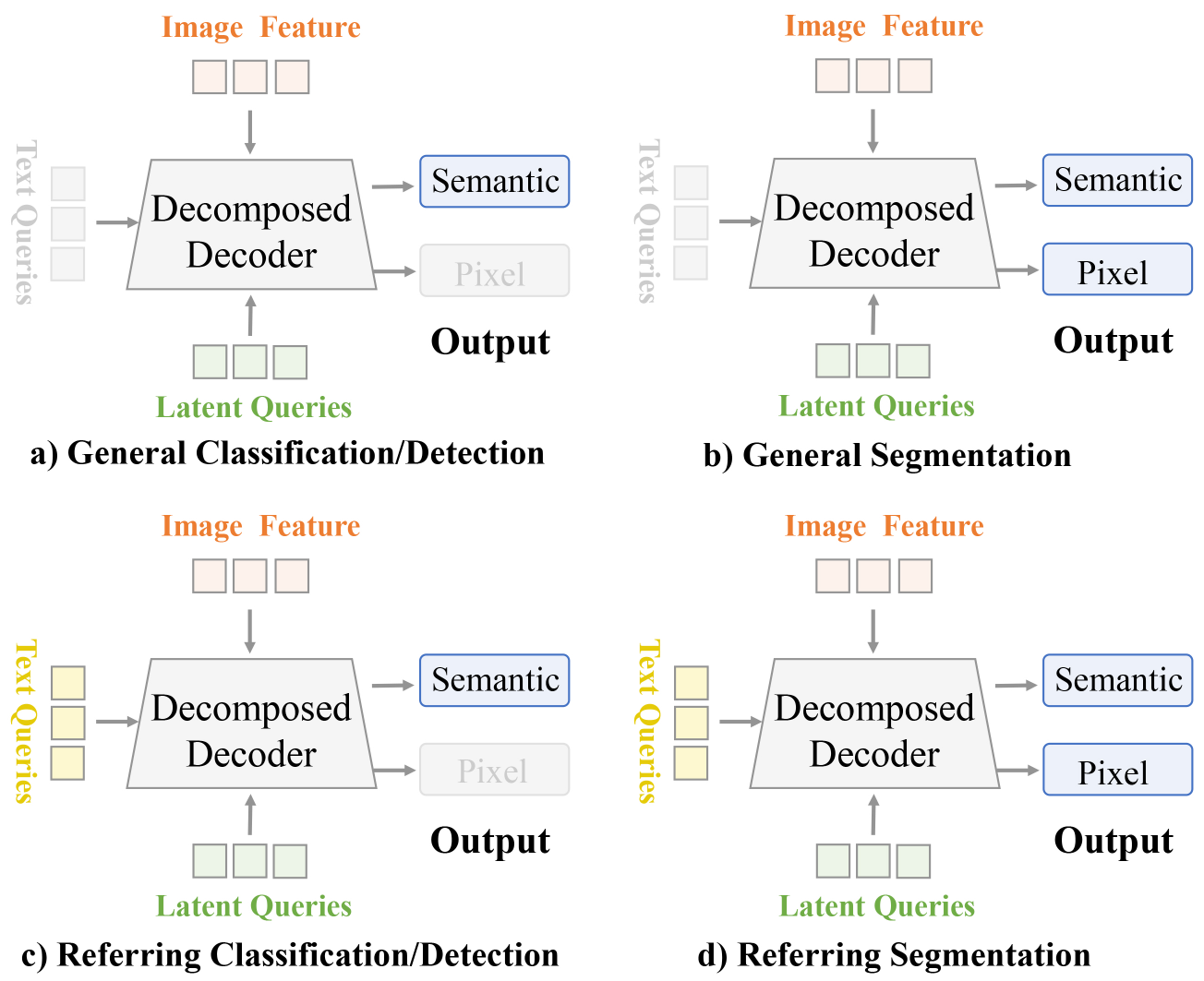}
		\caption{UniMed exhibits the capability to perform various medical image analysis tasks by dynamically combining input and output terminals. Specifically, include a) General classification/detection. b) General segmentation. c) Referring classification/detection. d) Referring segmentation. }
		\label{fig3}
	\end{figure}

	\textbf{General Classification/Detection (Fig. \ref{fig3} a)).}  General classification and detection rely on the input image for prediction, only leveraging visual features and general queues as input to the decomposed decoder. Through composed decoders, all annotations are tokenized, and the classification or detection task directly outputs the prediction results through the semantic path. Hence, expressed as: 
	\begin{equation}\label{eq6}
		[O_s] = \texttt{Dec} (V, (Q_g))
	\end{equation}
	
	\textbf{General Segmentation (Fig. \ref{fig3} b)).}  The input of its decomposed decoder is consistent with the Eq.\ref{eq2}. The image is encoded and make predictions by simultaneously generating pixel-level and semantic-level outcomes. The operation is as follows:
	\begin{equation}\label{eq7}
		[O_p, O_s] =  \texttt{Dec} (V, (Q_g)) 
	\end{equation} 
	
	\textbf{Referring Classification/Detection (Fig. \ref{fig3} c)).} The referring task requires a combination of visual features, text, and general queues (Eq. \ref{eq1}) to derive corresponding segmentation results. This enables clinical practice to flexibly obtain precise localization and diagnostic predictions of specified lesions by giving additional text prompts.
	\begin{equation}\label{eq8}
		[O_s] =  \texttt{Dec} (V, (Q_t, Q_g))  
	\end{equation} 
	
	\textbf{Referring Segmentation (Fig. \ref{fig3} d)).} It requires latent query and text query as input, so the formula is the same as Eq.\ref{eq5}. 
	\begin{equation}\label{eq9}
		[O_p, O_s] = \texttt{Dec} (V, (Q_t, Q_g)) 
	\end{equation}
	Compared with Eq. \ref{eq7}, the referring segmentation can be regarded general task with the language conditions.
	
	Through the combined arrangement of queues and outputs, UniMed can support a variety of medical imaging tasks (Fig. \ref{fig3}). This paper advocates for achieving unity through functional cohesion rather than interface specifications, thereby maximizing the shared utilization of common components across diverse tasks while preserving independence for each task.
	
	\subsection{Joint Representation Learning } 
	For medical image pre-training, relying solely on labeled data is far from enough, especially compared with the millions of data available in natural images. Hence, we delve into strategies for leveraging unlabeled data for training, aiming to bridge the disparity. Guided by this principle and drawing inspiration from the self-supervised contrastive learning paradigm, this work devises a joint training methodology, enabling the learning of labeled and unlabeled data simultaneously.

	\textit{\begin{table*}[!t] 
			\caption{Comparing our UniMed \textbf{fine-tuning} with the recent SOTA of \textbf{detection} task and general models outperforms all methods. "\textbf{Number}" indicates the best result, and "\underline{number}" indicates the suboptimal result.}
			\centering
			\renewcommand{\arraystretch}{1}
			\resizebox{0.9\textwidth}{!}{
				\begin{tabular}{c|cccccc|ccccc|c}
					\hline
					\multirow{2}*{Method}&STFT &Mask R-CNN &YOLOv8 &Trans VOD &DETR &Dyhead &Pix2Seq v2&	Unified-IO&GLIPv2 &Uni-Perceiver v2 &X-Decoder&Ours \\
					&\multicolumn{6}{c|}{Specific}&\multicolumn{5}{c|}{General-purpose}&Universal\\
					\hline
					SUN&\multirow{2}*{36.1}&\multirow{2}*{49.2}&\multirow{2}*{53.5}&\multirow{2}*{45.0}&\multirow{2}*{43.5}&\multirow{2}*{\underline{53.6}}&\multirow{2}*{48.8}&\multirow{2}*{50.2}&\multirow{2}*{53.4}&\multirow{2}*{51.1}&\multirow{2}*{51.6}&\multirow{2}*{\textbf{56.8} \tiny(+3.2)}\\
					(mAP)&&&&&&&&&&&&\\
					\hline
			\end{tabular}}
			\label{tabdetection}
	\end{table*}}
	
	\begin{table*}[!t] 
		\caption{Comparing our UniMed \textbf{fine-tuning} with the recent SOTA of \textbf{classification} task and general models outperforms all methods. "-" indicates that the model is not capable of handling a specific task. }
		\centering
		\renewcommand{\arraystretch}{1}
		\resizebox{0.9\textwidth}{!}{
			\begin{tabular}{c|cccccc|ccccc|c}
				\hline
				\multirow{2}*{Method}&ResNet&
				EfficientNet &CoAtNet &ViT-G/14 &SwinV2 &Model soups  &Pix2Seq v2&	Unified-IO&GLIPv2 &Uni-Perceiver v2 &X-Decoder&Ours \\
				&\multicolumn{6}{c|}{Specific}&\multicolumn{5}{c|}{General-purpose}&Universal\\
				\hline
				ColonCG&\multirow{2}*{84.4}&\multirow{2}*{88.4}&\multirow{2}*{88.1}&\multirow{2}*{89.0}&\multirow{2}*{88.9}&\multirow{2}*{\underline{89.1}}&\multirow{2}*{-}&\multirow{2}*{87.7}&\multirow{2}*{87.6}&\multirow{2}*{88.2}&\multirow{2}*{87.8}&\multirow{2}*{\textbf{90.8} \tiny(+1.7)}\\
				(mAcc)&&&&&&&&&&&&\\
				\hline
		\end{tabular}}
		\label{tabcls}
	\end{table*}
	
	\begin{table*}[!t] 
		\caption{Comparing our UniMed \textbf{fine-tuning} with the recent SOTA of \textbf{segmentation} task and general models outperforms all methods. "-" indicates that the model is not capable of handling a specific task. }
		\centering
		\renewcommand{\arraystretch}{0.9}
		\resizebox{\textwidth}{!}{
			\begin{tabular}{c|cccccccc|ccccc|c}
				\hline
				\multirow{2}*{Method}&UNet&
				PraNet &SANet &BoxPolyp &nnUNet &TransUNet  &Mask2Former&SegViT-V2&Pix2Seq v2&	Unified-IO&GLIPv2 &Uni-Perceiver v2 &X-Decoder&Ours \\
				&\multicolumn{8}{c|}{Specific}&\multicolumn{5}{c|}{General-purpose}&Universal\\
				\hline
				\multirow{2}*{CVC-ClinicDB}&\multirow{2}*{81.3}&\multirow{2}*{91.3}&\multirow{2}*{92.2}&\multirow{2}*{93.0}&\multirow{2}*{91.8}&\multirow{2}*{91.3}&\multirow{2}*{91.8}&\multirow{2}*{92.6}&\multirow{2}*{90.8}&\multirow{2}*{92.5}&\multirow{2}*{-}&\multirow{2}*{91.8}&\multirow{2}*{\underline{93.1}}&\multirow{2}*{\textbf{93.6} \tiny(+0.5)}\\
				&&&&&&&&&&&&&&\\
				\multirow{2}*{CVC-ColonDB}&\multirow{2}*{66.1}&\multirow{2}*{75.7}&\multirow{2}*{75.7}&\multirow{2}*{79.9}&\multirow{2}*{74.6}&\multirow{2}*{76.2}&\multirow{2}*{78.9}&\multirow{2}*{79.3}&\multirow{2}*{76.5}&\multirow{2}*{77.1}&\multirow{2}*{-}&\multirow{2}*{\underline{80.1}}&\multirow{2}*{78.2}&\multirow{2}*{\textbf{80.9} \tiny(+0.8)}\\
				&&&&&&&&&&&&&&\\
				\multirow{2}*{Kvasir-SEG}&\multirow{2}*{83.8}&\multirow{2}*{88.7}&\multirow{2}*{88.7}&\multirow{2}*{91.4}&\multirow{2}*{90.5}&\multirow{2}*{93.2}&\multirow{2}*{92.5}&\multirow{2}*{93.1}&\multirow{2}*{91.7}&\multirow{2}*{92.4}&\multirow{2}*{-}&\multirow{2}*{\underline{93.2}}&\multirow{2}*{91.6}&\multirow{2}*{\textbf{94.1} \tiny(+0.9)}\\
				&&&&&&&&&&&&&&\\
				\multirow{2}*{ETIS-LaribPolypDB}&\multirow{2}*{51.9}&\multirow{2}*{73.0}&\multirow{2}*{73.0}&\multirow{2}*{81.3}&\multirow{2}*{72.8}&\multirow{2}*{\underline{81.5}}&\multirow{2}*{80.5}&\multirow{2}*{80.1}&\multirow{2}*{75.2}&\multirow{2}*{79.2}&\multirow{2}*{-}&\multirow{2}*{80.6}&\multirow{2}*{78.1}&\multirow{2}*{\textbf{88.1} \tiny(+6.6)}\\
				&&&&&&&&&&&&&&\\
				EndoScene&\multirow{2}*{84.3}&\multirow{2}*{88.5}&\multirow{2}*{88.1}&\multirow{2}*{88.4}&\multirow{2}*{88.6}&\multirow{2}*{88.3}&\multirow{2}*{88.6}&\multirow{2}*{89.7}&\multirow{2}*{87.4}&\multirow{2}*{87.6}&\multirow{2}*{-}&\multirow{2}*{88.5}&\multirow{2}*{\underline{89.8}}&\multirow{2}*{\textbf{91.8} \tiny(+2.0)}\\
				(Dice)&&&&&&&&&&&&&&\\
				\hline
		\end{tabular}}
		\label{tabseg}
	\end{table*}
	
	\textbf{Pipeline.} As for labeled data, the corresponding labels undergo standard transformations as outlined in Sec. \ref{sec3.2}. These labels are categorized into semantic and pixel types, and training is conducted utilizing both semantic loss and pixel loss. For unlabeled data, follow the dense contrastive learning strategy \cite{wang2021dense}.  For each image, we generate two sets of random views via data augmentation and feed them into the encoder to obtain two sets of features. These features are separately passed through downstream dense projection heads, and the same encoder is trained by computing the contrastive learning loss between the two sets of features. During training, we adopt an exponential moving average to update the parameters and retain the encoder part after training is completed, discarding the dense header. During inference, task predictions are executed through the adaptable combination of visual, text, and latent queue input terminals, alongside semantic and pixel output terminals. The total training loss can be expressed as a combination of these:
	
	\begin{equation}
		L_{total} = \begin{matrix} \underbrace{  L_{s} + L_{p}} \\
			L_{label} \end{matrix} + \lambda  \begin{matrix} \underbrace{ L_{c} + L_{dc}} \\
			L_{unlabel} \end{matrix}
	\end{equation} 
	
	Where $\lambda$ acts as a weight to balance the two terms. The semantic output $L_{s}$ is the cross entropy loss, the pixel output $L_{e}$ includes the binary cross entropy loss and the dice loss, and the unlabeled learning loss includes the contrast $L_{c}$ and the dense contrast loss $L_{dc}$.
	
	
	
	

	\section{Experimental Results}
	
	\begin{flushright}
		\begin{table*}[!t] 
			\caption{Comparing the \textbf{zero-shot} and \textbf{100-shot} performance of our UniMed to the recent universal model, it outperforms all methods. "-" indicates that the model is not capable of handling a specific task. }
			\centering
			\renewcommand{\arraystretch}{1.1}
			\resizebox{0.8\textwidth}{!}{
				\begin{tabular}{cc|c|c|ccccc}
					\hline
					~&\multirow{3}*{Method}&Detection &Classification &\multicolumn{5}{c}{Segmentation}\\
					~&~&(mAP)&(mAcc)&\multicolumn{5}{c}{(Dice)}\\
					~&~&SUN&ColonCG& \makecell{CVC-ClinicDB}&\makecell{CVC-ColonDB}&\makecell{Kvasir-SEG}& \makecell{ETIS-LaribPolypDB} &\makecell{EndoScene}\\
					\hline
					\multirow{6}*{Zero-shot}&GLIPv2 \cite{zhang2022glipv2} &20.6&48.4&\multicolumn{5}{c}{-}\\
					~& MM-G-T \cite{zhao2024open}&26.4&53.8&\multicolumn{5}{c}{-}\\
					~&Uni-Perceiver v2 \cite{li2023uni}&\underline{33.1}&\underline{55.6}&48.3&33.6&\underline{57.8}&34.7&48.3\\
					~&X-Decoder \cite{zou2023generalized}&31.6&54.7&\underline{52.7}&\underline{35.4}&56.5&\underline{36.1}&\underline{50.8}\\
					\cline{2-9}
					~&  ~&\textbf{39.8}& \textbf{62.5}& \textbf{54.6}& \textbf{38.9}& \textbf{60.8}& \textbf{43.3}& \textbf{55.7}\\
					~&  \multirow{-2}*{UniMed}& (+6.7)& (+6.9)& (+1.9)& (+3.5)& (+3.0)& (+7.2)& (+4.9)\\
					\hline  
					\multirow{6}*{100-shot}&GLIPv2  &41.8&83.4&\multicolumn{5}{c}{-}\\
					~& MM-G-T &\underline{46.6}&\underline{85.5}&\multicolumn{5}{c}{-}\\
					~&Uni-Perceiver v2 &45.2&84.4&81.5&\underline{87.4}&86.3&72.6&84.6\\
					~&X-Decoder &46.3&\underline{85.5}&\underline{87.1}&\underline{72.9}&71.1&\underline{73.8}&\underline{84.9}\\
					\cline{2-9}
					~&  ~& \textbf{49.8}& \textbf{88.4}& \textbf{88.5}& \textbf{75.2}& \textbf{89.1}& \textbf{77.9}& \textbf{87.3}\\
					~&   \multirow{-2}*{UniMed}& (+3.2)& (+2.9)& (+1.4)& (+2.3)& (+1.7)& (+4.1)& (+2.4)\\
					\hline 
			\end{tabular}}
			\label{tab3}
		\end{table*}
	\end{flushright}
	
	\subsection{Tasks and Datasets}
	
	This study takes the endoscopic modality of medical images as an example and conducts a comprehensive investigation by collecting datasets from various research groups worldwide. The database established contains 12 datasets and covers all 3 tasks. The unlabeled datasets are endoscopic videos that are difficult to label, Colonoscopic \cite{mesejo2016computer}, Hyper-Kvasir \cite{borgli2020hyperkvasir}, Kvasir-Capsule \cite{smedsrud2021kvasir}, LDPolypVideo \cite{ma2021ldpolypvideo}, and ColonVideo (private, from Jiangsu Provincial People's Hospital). For detection, evaluation is performed on the SUN \cite{misawa2021development} colonoscopy public dataset, the largest benchmark for polyp detection. The classification task utilizes the ColonCG (Private, from Jiangsu Provincial People's Hospital), which is the most comprehensive dataset for colon disease classification, including five categories: normal, polyp, adenoma, cancer, and ulcerative colitis. Segmentation tasks are evaluated on the common public polyp segmentation datasets: CVC-ClinicDB \cite{bernal2015wm}, CVC-ColonDB \cite{tajbakhsh2015automated}, Kvasir-SEG \cite{jha2020kvasir}, ETIS-LaribPolypDB \cite{silva2014toward}, and EndoScene \cite{vazquez2017benchmark}. Evaluation metrics include mean average precision (mAP), mean accuracy (mAcc), and Dice corresponding to the three tasks respectively.
	

	\subsection{Implementation Details}
	We employ Focal-T \cite{yang2022focal} as the backbone of the visual encoder, utilizing a transformer text encoder with causal masking \cite{radford2021learning} as the language encoder. For training, the AdamW \cite{loshchilov2017decoupled} optimizer with a base learning rate set to 1e-4, a weight decay of 0.05, and a linear decay learning rate scheduler are applied. The training procedure spans 50 epochs with a batch size of 8. A total of two data loaders are used: one for labeled data and another for unlabeled data, maintaining a sampling ratio of 1:1. The final loss function comprises both supervised and unsupervised components, with a ratio of 10:1 to balance their contributions effectively. During the fine-tuning process, the model's input and output are controlled through configuration files, allowing for customized settings tailored to specific task requirements.
	
	\begin{figure}[tb]
		\centering
		\includegraphics[width=\columnwidth]{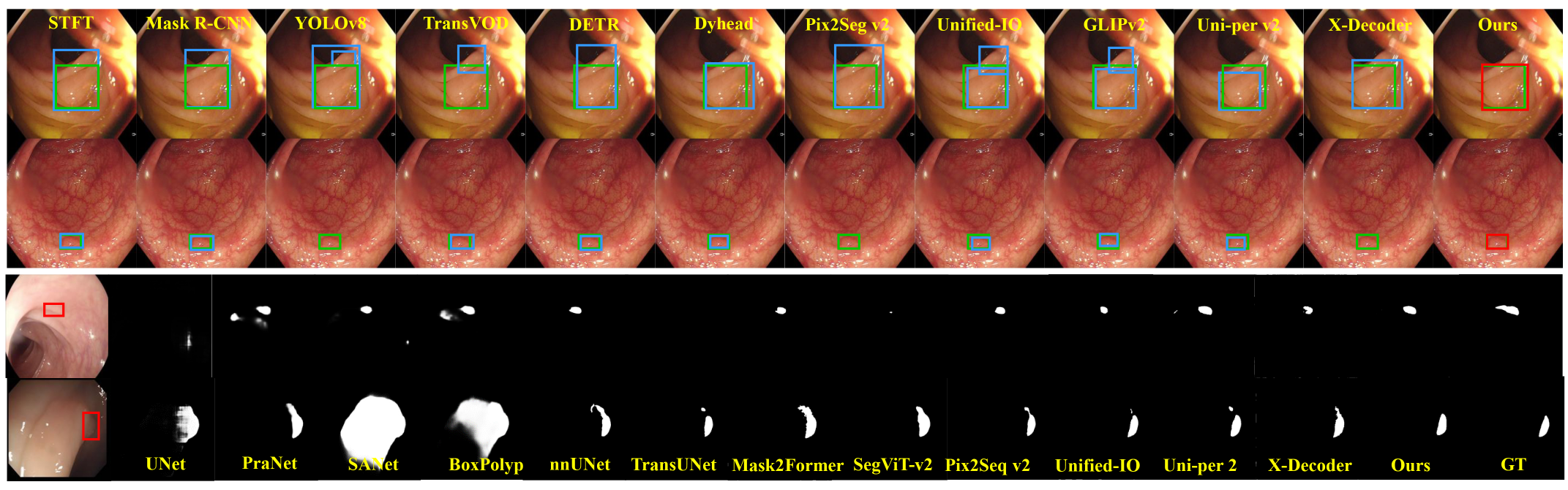}
		\caption{Visualization results on detection and segmentation tasks compared with other methods.
		}
		\label{figvis}
	\end{figure}
	
	\subsection{Task-specific Fine-tuning}
	
	UniMed undergoes comparison with existing specialized and universal methods across various tasks. Fine-tuning is performed on 8 datasets for 3 common medical image analysis downstream tasks, with performance reported in Table \ref{tabdetection}, \ref{tabcls}, \ref{tabseg}. The analysis yields the following observations.
	
	\textbf{UniMed v.s. Specialized Detection Methods.} In a comprehensive evaluation, UniMed is compared with various state-of-the-art methods (Table \ref{tabdetection}), including polyp detection \cite{wu2021multi}, two-stage \cite{he2017mask}, single-stage \cite{jocher2020ultralytics}, and transformer-based detection methods \cite{zhou2022transvod, carion2020end, dai2021dynamic}.  UniMed surpasses all specialized detection methods (Fig. \ref{figvis}), achieving the highest mean Average Precision (mAP) score of 56.8\%, outperforming the next-best result by a significant margin of 3.2\%.
	
	\textbf{UniMed v.s. Specialized Classification Methods.} A total of six methods are compared (Table \ref{tabcls}), encompassing CNN-based \cite{he2016deep, tan2019efficientnet} , Transformer-based \cite{zhai2022scaling, liu2022swin}, and CNN-Transformer hybrid approaches  \cite{dai2021coatnet, wortsman2022model}. UniMed emerges as the top performer, achieving the highest classification results with an average accuracy of 90.8\% across five categories. Notably, among these methods, Model Soups achieves commendable results, trailing behind the proposed UniMed model by on 1.7 points. This observation underscores the effectiveness of averaging multiple weights in Model soups, contributing to its competitive performance despite being sub-optimal.
	
	\textbf{UniMed v.s. Specialized Segmentation Methods.} When comparing UniMed with state-of-the-art methods in polyp segmentation \cite{falk2019u, fan2020pranet, wei2021shallow, wei2022boxpolyp} and semantic segmentation  (Table \ref{tabseg}) \cite{isensee2021nnu, chen2021transunet, cheng2022masked, zhang2023segvitv2}, UniMed consistently achieves the best segmentation results (Fig. \ref{figvis}) across datasets with varying levels of segmentation difficulty. Specifically, UniMed outperforms suboptimal methods by 0.6, 1.0, 0.9, 6.6, and 2.1 points on the CVC-ClinicDB, CVC-ColonDB, Kvasir-SEG, ETIS-LaribPolypDB, EndoScene datasets, respectively. These results underscore the robustness of UniMed's performance across diverse datasets, demonstrating its effectiveness in tackling segmentation challenges across different medical imaging scenarios.
	
	\textbf{UniMed v.s. Generalist Methods.} Most notably, the proposed end-to-end architecture outperforms other general models across various medical tasks (Table \ref{tabdetection}, \ref{tabcls}, \ref{tabseg}). General methods \cite{chen2022unified, lu2022unified, zhang2022glipv2, li2023uni, zou2023generalized} tend to exhibit higher overall performance compared to specific methods, suggesting that the interaction of data and tasks fosters enhanced model learning. Additionally, models striving for high understanding performance often demonstrate lower localization performance (e.g., Uni-Perceiver v2), as it is not trivial to merge semantic and visual understanding into a single model. Similarly (Fig. \ref{figvis}), visual comparison results with other methods clearly show that our method has higher accuracy in boundary segmentation and localization detection.

	\subsection{Zero-Shot and 100-Shot Transfer} 
	UniMed is pre-trained, requiring only zero or a small number of parameters before its application to various downstream tasks. Thus, we assessed the model's transferability to other tasks in both zero-shot and 100-shot settings.
	
	\textbf{Zero-shot Transfer.} Experimental results demonstrate compelling evidence of UniMed's substantial zero-shot capability in the medical field compared to other general methods (Table \ref{tab3}). This suggests that UniMed can be readily applied to various tasks without further adjustments. Moreover, UniMed surpassed sub-optimal results by 6.7\% and 6.9\% in detection and classification tasks, respectively. Additionally, the performance improvement in segmentation tasks even outperforms other methods by up to 7.2\%.
	
	\textbf{100-shot Transfer.} UniMed also demonstrates the superior overall performance of strong 100-shot on medical image analysis tasks (Table \ref{tab3}). In some instances, it even rivals fully supervised models trained with full-scale data, exemplified by the SUN dataset (100-shot AP 49.8\%, compared to Mask RCNN's 49.2\%). In particular, comparing it with the X-decoder model, it can be seen that its performance is exceeded in both zero-shot and 100-shot cases. This underscores the key role of large amounts of unsupervised data in feature generalization. 
	
	\begin{figure}[tb]
		\centering
		\includegraphics[width=\columnwidth]{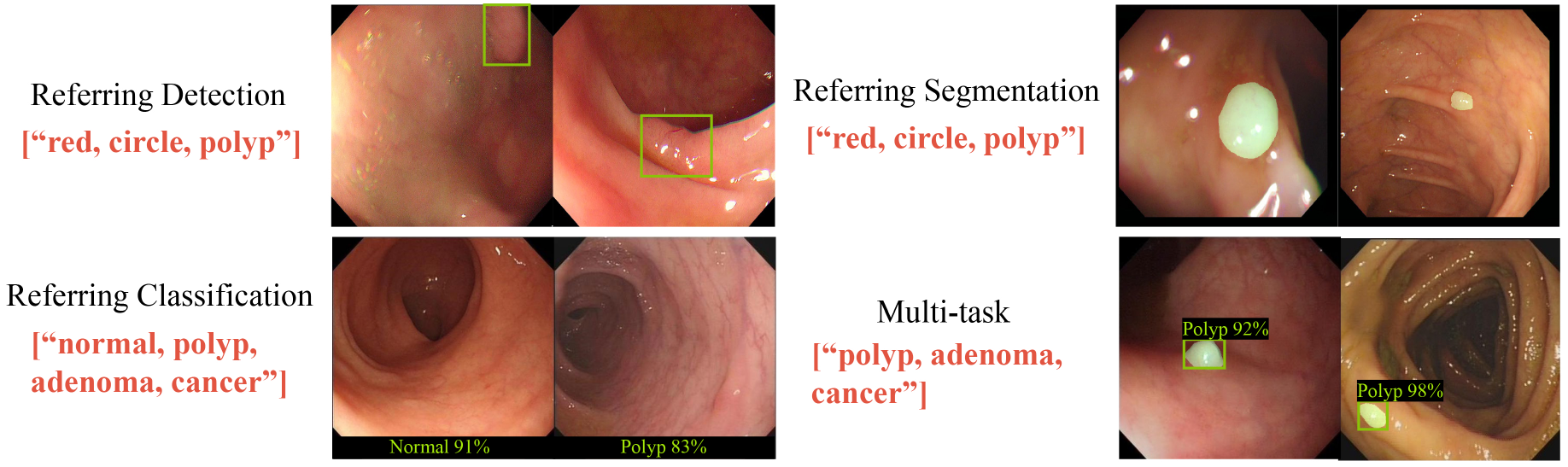}
		\caption{Qualitative results demonstrate UniMed's ability to support referring tasks and help clinically obtain specified predictions.
		}
		\label{fig4}
	\end{figure}
	
	\subsection{Task Composition}
	As mentioned earlier, UniMed boasts a unique advantage of task interaction, enabling both single and joint-task reasoning. This distinctive capability enhances the model's practicality in real clinical scenarios, particularly in customizing referrals and tasks. In Fig. \ref{fig4}, visualization results of single and joint task inference without architectural changes are showcased. For instance, when provided with a set of referrals such as ["polyp, adenoma, cancer"], UniMed seamlessly delivers both pixel-level and semantic-level predictions.
	
	\subsection{Ablation Study}
	Ablation studies are performed to analyze the architecture of UniMed, and all experiments are tested on the segmentation task.
	
	\textbf{Backbone.} Increasing the size of the backbone network indeed leads to performance improvements, as evidenced in Table \ref{tab4}. As the depth and embedding dimensions expand the visual encoder, performance improves across the board. This suggests that more powerful feature extractors facilitate prediction for downstream tasks.
	
	\begin{table}[!t] 
		\caption{Ablation of the \textbf{backbone} network of visual encoders. "\textbf{Number}" indicates the best result.}
		\centering
		\renewcommand{\arraystretch}{1.2}
		\resizebox{\columnwidth}{!}{
			\begin{tabular}{c|ccccc}
				\hline
				Backbone  & CVC-ClinicDB &  CVC-ColonDB & Kvasir-SEG & ETIS-LaribPolypDB & EndoScene\\
				\hline 
				ViT-T&93.1&78.1&92.5&84.3&91.2\\
				ViT-L&\textbf{93.6} \tiny(+0.5) &\textbf{80.9} \tiny(+2.8)&\textbf{94.1} \tiny(+1.6)&\textbf{88.1} \tiny(+3.8)&\textbf{91.8} \tiny(+0.6)\\
				\hline 
		\end{tabular}}
		\label{tab4}
	\end{table}
	
	\begin{table}[!t] 
		\caption{Ablation of \textbf{collaboration and interference between tasks} by removing one task at a time. In the brackets are the gaps to the “All Tasks” counterpart. }
		\centering
		\renewcommand{\arraystretch}{1.2}
		\resizebox{\columnwidth}{!}{
			\begin{tabular}{c|ccccc}
				\hline
				Task & CVC-ClinicDB  &CVC-ColonDB & Kvasir-SEG & ETIS-LaribPolypDB &EndoScene\\
				\hline 
				All Tasks &\textbf{93.6}&80.9&\textbf{94.1}&88.1&91.8\\
				-Detection&93.3 \tiny(-0.3)&79.8 \tiny(-1.1)&93.6 \tiny(-0.5)&86.7 \tiny(-1.4)&\textbf{92.0} \tiny(+0.2)\\
				-Classification&92.9 \tiny(-0.7) &\textbf{81.4} \tiny(+0.5)&93.6 \tiny(-0.5)&\textbf{87.8} \tiny(-0.3)&91.0 \tiny(-0.8)\\
				Single Segmentation &92.5 \tiny(-1.1)&79.5 \tiny(-1.4)&93.4 \tiny(-0.7)&87.6 \tiny(-1.5)&89.5 \tiny(-1.3)\\
				\hline 
		\end{tabular}}
		\label{tab5}
	\end{table}
	
	\textbf{Task Collaboration.} To explore the relationship between task collaboration and interference, by eliminating individual tasks based on all tasks (Table \ref{tab5}), the following finding was drawn: 1) Learning multiple tasks together is more effective than focusing on a single task alone. 2) Detection tasks positively influence segmentation tasks, whereas classification tasks exhibit suboptimal performance on certain datasets. This discrepancy may be attributed to the interference caused by dividing images into multiple categories, particularly when dealing with challenging data that is difficult to classify accurately.
	
	\begin{table}[!t] 
		\caption{Ablation of \textbf{weights between labeled and unlabeled losses} in Eq. \ref{eq7}.  "\underline{number}" indicates the suboptimal result.}
		\centering
		\renewcommand{\arraystretch}{1.2}
		\resizebox{\columnwidth}{!}{
			\begin{tabular}{c|ccccc}
				\hline
				Loss weights \textbf{$\lambda$} &  CVC-ClinicDB&  CVC-ColonDB  &  Kvasir-SEG &  ETIS-LaribPolypDB  & EndoScene \\
				\hline 
				1&92.5&79.5&93.7&85.5&89.5\\
				0.75&91.8&79.4&\underline{94.0}&84.3&\underline{90.9}\\
				0.5&\underline{93.2}&\underline{80.6}&93.6&\underline{85.9}&\textbf{91.8}\\
				0.1&\textbf{93.6}&\textbf{80.9}&\textbf{94.1}&\textbf{88.1}&\textbf{91.8}\\
				0&92.8  &78.7 &92.2&82.5 &90.6\\
				\hline 
		\end{tabular}}
		\label{tab6}
	\end{table}
	
	\textbf{Balance between Labeled and Unlabeled Losses.} Table \ref{tab6} demonstrates that the best performance is achieved when the unsupervised loss is set to 0.1, suggesting that the supervised component holds greater significance in the overall training process. When the ratio of unsupervised to supervised losses is 1:1, there is a drop in performance. These observations indicate that 1) it is necessary to introduce a joint expression learning strategy, where unlabeled data helps the model learn more general features. 2) The supervised component effectively drives model learning, while the unsupervised component serves as a "regularization" mechanism, guiding the model to acquire more robust knowledge. 
	
	\begin{table}[!t] 
		\scriptsize
		\setlength{\belowcaptionskip}{4pt}
		\caption{Ablation of the \textbf{sampling ratio} of labeled data and unlabeled data in data loading. "\underline{number}" indicates the suboptimal result.}
		\centering
		\renewcommand{\arraystretch}{1.2}
		\resizebox{\columnwidth}{!}{
			\begin{tabular}{c|ccccc}
				\hline
				Sampling ratio \textbf{$\lambda$} &CVC-ClinicDB & CVC-ColonDB  &  Kvasir-SEG  &  ETIS-LaribPolypDB & EndoScene \\
				\hline 
				0.5:1&93.2&79.6&92.7&85.8&91.2\\ 1:1&\underline{93.6}&\underline{80.9}&\textbf{94.1}&\textbf{88.1}&\textbf{91.8}\\
				1:2&\textbf{93.8}&\textbf{81.2}&\underline{93.4}&\underline{86.6}&\underline{91.5}\\
				\hline 
		\end{tabular}}
		\label{tab7}
	\end{table}
	
	\textbf{Balance between Labeled and Unlabeled Data.} Table \ref{tab7} shows that during experimental loading, sampling ratios between labeled and unlabeled data below 1 perform better. That is, the sampling ratio of unlabeled data should be higher. This is because, in medical data, the amount of labeled data is very small, while unlabeled data can significantly increase the diversity of samples. Based on the above findings, this paper adopts a 1:1 ratio setting.
	
	\section{Conclusion}
	We present UniMed, the first general-purpose architecture designed for comprehensive medical images to support tasks at all levels, including image, region, and pixel levels. Unlike the current universal models that usually involve multiple task-specific branches or heads and rely on cumbersome multi-stage pre-training processes. The innovative coupling design of the decmposed-composed decoders unifies the input and output space, enabling flexible model combinations to support various task interactions seamlessly. The joint representation learning strategy demonstrates how to effectively train models in a single stage without additional modules. Our approach addresses the challenges of annotation diversity and underutilization of unlabeled samples in medical data, achieving state-of-the-art performance in fine-tuning, zero-shot, and 100-shot scenarios on eight datasets. Overall, we believe that UniMed has significant advantages in real-world clinical applications due to its versatility, transferability, and flexibility. 

\bibliography{aaai25}

\end{document}